\def\year{2019}\relax
\documentclass[letterpaper]{article} 
\usepackage{aaai19}  
\usepackage{times}  
\usepackage{helvet}  
\usepackage{courier}  
\usepackage{url}  
\usepackage{graphicx}  
\usepackage{multirow}
\usepackage{amsmath}
\usepackage{amssymb}
\usepackage{enumerate}
\usepackage{graphicx}
\usepackage{verbatim}
\usepackage{graphics}

\begin{document}
%
\title{Dynamic Compositionality in Recursive Neural Networks with Structure-aware Tag Representations}
\author{$^\dagger$Taeuk Kim, $^\dagger$Jihun Choi, $^\ddagger$Daniel Edmiston, $^\dagger$Sanghwan Bae, $^\dagger$Sang-goo Lee\\
$^\dagger$Department of Computer Science and Engineering, Seoul National University, Seoul, Korea\\
\{taeuk, jhchoi, sanghwan, sglee\}@europa.snu.ac.kr\\
$^\ddagger$Department of Linguistics, University of Chicago, Chicago, IL, USA\\
danedmiston@uchicago.edu
}

\maketitle
\begin{abstract}
Most existing recursive neural network (RvNN) architectures utilize only the structure of parse trees, ignoring syntactic tags which are provided as by-products of parsing. 
We present a novel RvNN architecture that can provide dynamic compositionality by considering comprehensive syntactic information derived from both the structure and linguistic tags. 
Specifically, we introduce a structure-aware tag representation constructed by a separate tag-level tree-LSTM. 
With this, we can control the composition function of the existing word-level tree-LSTM by augmenting the representation as a supplementary input to the gate functions of the tree-LSTM. 
In extensive experiments, we show that models built upon the proposed architecture obtain superior or competitive performance on several sentence-level tasks such as sentiment analysis and natural language inference when compared against previous tree-structured models and other sophisticated neural models.
\end{abstract}

\section{Introduction}
One of the most fundamental topics in natural language processing is how best to derive high-level representations from constituent parts, as natural language meanings are a function of their constituent parts. How best to construct a sentence representation from distributed word embeddings is an example domain of this larger issue. 
Even though sequential neural models such as recurrent neural networks (RNN) \cite{elman1990finding} and their variants including Long Short-Term Memory (LSTM) \cite{hochreiter1997long} and Gated Recurrent Unit (GRU) \cite{cho2014learning} have become the de-facto standard for condensing sentence-level information from a sequence of words into a fixed vector, there have been many lines of research towards better sentence representation using other neural architectures, e.g. convolutional neural networks (CNN) \cite{kim2014convolutional} or self-attention based models \cite{shen2018reinforced}.

From a linguistic point of view, the underlying tree structure---as expressed by its constituency and dependency trees---of a sentence is an integral part of its meaning. 
Inspired by this fact, some recursive neural network (RvNN\footnote{To avoid confusion, we call recursive neural networks (or tree-structured NNs) RvNNs to distinguish them from recurrent neural networks RNNs, following the convention of some previous works.}) models are designed to reflect the syntactic tree structure, achieving impressive results on several sentence-level tasks such as sentiment analysis \cite{socher2012semantic,socher2013recursive}, machine translation \cite{yang2017towards}, natural language inference \cite{bowman2016fast}, and discourse relation classification \cite{wang2017tag}. 

However, some recent works have \cite{yogatama2017learning,choi2018learning} proposed latent tree models, which learn to construct task-specific tree structures without explicit supervision, bringing into question the value of linguistically-motivated recursive neural models.
Witnessing the surprising performance of the latent tree models on some sentence-level tasks, there arises a natural question: \textit{Are linguistic tree structures the optimal way of composing sentence representations for NLP tasks?}

In this paper, we demonstrate that linguistic priors are in fact useful for devising effective neural models for sentence representations, showing that our novel architecture based on constituency trees and their tag\footnote{In this work, we refer to both part-of-speech (POS) tags (e.g. DT-determiner, JJ-adjective) for words and phrase-level tags (e.g. NP-noun phrase, VP-verb phrase) simply as `tags'.} information obtains superior performance on several sentence-level tasks, including sentiment analysis and natural language inference. 

A chief novelty of our approach is that we introduce a small separate tag-level tree-LSTM to control the composition function of the existing word-level tree-LSTM, which is in charge of extracting helpful syntactic signals for meaningful semantic composition of constituents by considering both the structures and linguistic tags of constituency trees simultaneously. 
In addition, we demonstrate that applying a typical LSTM to preprocess the leaf nodes of a tree-LSTM greatly improves the performance of the tree models. 
Moreover, we propose a clustered tag set to replace the existing tags on the assumption that the original syntactic tags are too fined-grained to be useful in neural models.

In short, our contributions in this work are as follows:
\begin{itemize}
    \item We propose a new linguistically-motivated neural model which generates high-quality sentence representations by considering all the information extracted from constituency parse trees. 
    \item In addition, we demonstrate the superiority of the proposed models achieving new state-of-the-art performance within the similar model class on 4 out of 5 sentence classification benchmarks, as well as showing competitive results compared to other types of neural models.
    \item We empirically show that another key point to the success of tree-structured models is to contextualize input word embeddings so that the corresponding input for each word in a sentence can better reflect the meaning of the whole sentence.
    
\end{itemize}

\section{Related Work}
Recursive neural networks (RvNN) are a kind of neural architecture which model sentences by exploiting syntactic structure.
While earlier RvNN models proposed utilizing diverse composition functions, including feed-forward neural networks \cite{socher2011parsing}, matrix-vector multiplication \cite{socher2012semantic}, and tensor computation \cite{socher2013recursive}, tree-LSTMs \cite{tai2015improved} remain the standard for several sentence-level tasks.

Even though classic RvNNs have demonstrated superior performance on a variety of tasks, their inflexibility, i.e. their inability to handle \textit{dynamic compositionality} for different syntactic configurations, is a considerable weakness. 
For instance, it would be desirable if our model could distinguish e.g. adjective-noun composition from that of verb-noun or preposition-noun composition, as models failing to make such a distinction ignore real-world syntactic considerations such as `-arity' of function words (i.e. types), and the adjunct/argument distinction.

To enable dynamic compositionality in recursive neural networks, many previous works \cite{hashimoto2013simple,dong2014adaptive,qian2015learning,wang2017tag,liu2017dynamic,huang2017encoding,teng2017head} have proposed various methods. 

One main direction of research leverages tag information, which is produced as a by-product of parsing.
In detail, \citeauthor{qian2015learning} (\citeyear{qian2015learning}) suggested TG-RNN, a model employing different composition functions according to POS tags, and TE-RNN/TE-RNTN, models which leverage tag embeddings as additional inputs for the existing tree-structured models.
Despite the novelty of utilizing tag information, the explosion of the number of parameters (in case of the TG-RNN) and the limited performance of the original models (in case of the TE-RNN/TE-RNTN) have prevented these models from being widely adopted.
Meanwhile, \citeauthor{wang2017tag} (\citeyear{wang2017tag}) and \citeauthor{huang2017encoding} (\citeyear{huang2017encoding}) proposed models based on a tree-LSTM which also uses the tag vectors to control the gate functions of the tree-LSTM. 
In spite of their impressive results, there is a limitation that the trained tag embeddings are too simple to reflect the rich information which tags provide in different syntactic structures. To alleviate this problem, we introduce structure-aware tag representations in the next section.

Another way of building dynamic compositionality into RvNNs is to take advantage of a meta-network (or hyper-network).
Inspired by recent works on dynamic parameter prediction, DC-TreeLSTMs \cite{liu2017dynamic} dynamically create the parameters for compositional functions in a tree-LSTM. Specifically, the model has two separate tree-LSTM networks whose architectures are similar, but the smaller of the two is utilized to calculate the weights of the bigger one. A possible problem for this model is that it may be easy to be trained such that the role of each tree-LSTM is ambiguous, as they share the same input, i.e. word information. Therefore, we design two disentangled tree-LSTMs in our model so that one focuses on extracting useful features from only syntactic information while the other composes semantic units with the aid of the features. Furthermore, our model reduces the complexity of computation by utilizing typical tree-LSTM frameworks instead of computing the weights for each example.

Finally, some recent works \cite{yogatama2017learning,choi2018learning} have proposed latent tree-structured models that learn how to formulate tree structures from only sequences of tokens, without the aid of syntactic trees or linguistic information. 
The latent tree models have the advantage of being able to find the optimized task-specific order of composition rather than a sequential or syntactic one. 
In experiments, we compare our model with not only syntactic tree-based models but also latent tree models, demonstrating that modeling with explicit linguistic knowledge can be an attractive option.

\section{Model}

In this section, we introduce a novel RvNN architecture, called \textbf{SATA Tree-LSTM}\footnote{The implementation of our model and supplemental materials are available at https://github.com/galsang/SATA-Tree-LSTM.} (\textbf{S}tructure-\textbf{A}ware \textbf{T}ag \textbf{A}ugmented \textbf{Tree-LSTM}). 
This model is similar to typical Tree-LSTMs, but provides dynamic compositionality by augmenting a separate tag-level tree-LSTM which produces structure-aware tag representations for each node in a tree. 
In other words, our model has two independent tree-structured modules based on the same constituency tree, one of which (word-level tree-LSTM) is responsible for constructing sentence representations given a sequence of words as usual, while the other (tag-level tree-LSTM) provides supplementary syntactic information to the former.

In section 3.1, we first review tree-LSTM architectures. 
Then in section 3.2, we introduce a tag-level tree-LSTM and structure-aware tag representations.
In section 3.3, we discuss an additional technique to boost the performance of tree-structured models, and in section 3.4, we describe the entire architecture of our model in detail.

\subsection{Tree-LSTM}
The LSTM \cite{hochreiter1997long} architecture was first introduced as an extension of the RNN architecture to mitigate the vanishing and exploding gradient problems. In addition, several works have discovered that applying the LSTM cell into tree structures can be an effective means of modeling sentence representations.

To be formal, the composition function of the cell in a tree-LSTM can be formulated as follows:
\begin{equation} \label{eq:1}
    \begin{bmatrix}
        \mathbf{i} \\
        \mathbf{f}_l \\
        \mathbf{f}_r \\ 
        \mathbf{o} \\
        \mathbf{g}
    \end{bmatrix}
    = 
    \begin{bmatrix}
         \sigma \\
         \sigma \\
         \sigma \\ 
         \sigma \\
         \tanh 
    \end{bmatrix}
    \Bigg( \mathbf{W}
    \begin{bmatrix}
        \mathbf{h}_l \\
        \mathbf{h}_r \\
    \end{bmatrix}
    + \mathbf{b} \Bigg)
\end{equation}
\begin{equation} \label{eq:2}
    \mathbf{c} = \mathbf{f}_l \odot \mathbf{c}_l + \mathbf{f}_r \odot \mathbf{c}_r + \mathbf{i} \odot \mathbf{g}\\ 
\end{equation}
\begin{equation} \label{eq:3}
    \mathbf{h} = \mathbf{o} \odot \tanh{\left(\mathbf{c}\right)}
\end{equation}

\noindent where $\mathbf{h}, \mathbf{c} \in\mathbb{R}^{d}$ indicate the hidden state and cell state of the LSTM cell, and $\mathbf{h}_l, \mathbf{h}_r, \mathbf{c}_l, \mathbf{c}_r \in\mathbb{R}^{d}$ the hidden states and cell states of a left and right child. 
$\mathbf{g} \in\mathbb{R}^{d}$ is the newly composed input for the cell and $\mathbf{i}, \mathbf{f}_{l}, \mathbf{f}_{r}, \mathbf{o} \in\mathbb{R}^{d}$ represent an input gate, two forget gates (left, right), and an output gate respectively. 
$\mathbf{W} \in\mathbb{R}^{5d\times2d}$ and $\mathbf{b} \in\mathbb{R}^{5d}$ are trainable parameters. 
$\sigma$ corresponds to the sigmoid function, $\tanh$ to the hyperbolic tangent, and $\odot$ to element-wise multiplication. 

Note the equations assume that there are only two children for each node, i.e. binary or binarized trees, following the standard in the literature. While RvNN models can be constructed on any tree structure, in this work we only consider constituency trees as inputs.

In spite of the obvious upside that recursive models have in being so flexible, they are known for being difficult to fully utilize with batch computations as compared to other neural architectures because of the diversity of structure found across sentences. 
To alleviate this problem, \citeauthor{bowman2016fast} (\citeyear{bowman2016fast}) proposed the SPINN model, which brings a shift-reduce algorithm to the tree-LSTM.
As SPINN simplifies the process of constructing a tree into only two operations, i.e. shift and reduce, it can support more effective parallel computations while enjoying the advantages of tree structures.
For efficiency, our model also starts from our own SPINN re-implementation, whose function is exactly the same as that of the tree-LSTM.

\subsection{Structure-aware Tag Representation}

\begin{figure}[!t]
\centering
\includegraphics[width=0.9\columnwidth]{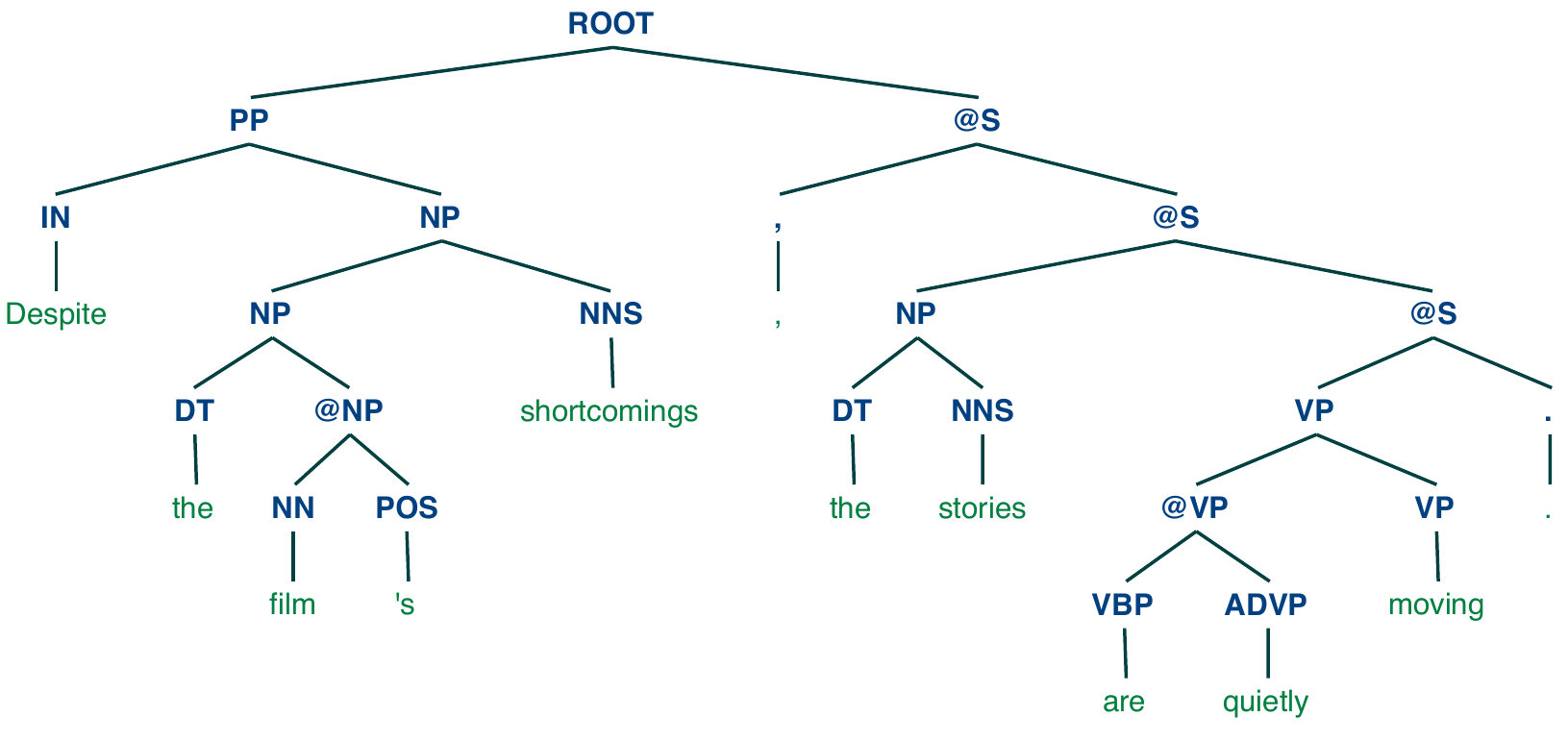}
\caption{A constituency tree example from Stanford Sentiment Treebank.}
\label{fig:figure1}
\end{figure}

In most previous works using linguistic tag information \cite{qian2015learning,wang2017tag,huang2017encoding}, tags are usually represented as simple low-dimensional dense vectors, similar to word embeddings. This approach seems reasonable in the case of POS tags that are attached to the corresponding words, but phrase-level constituent tags (e.g. NP, VP, ADJP) vary greatly in size and shape, making them less amenable to uniform treatment. For instance, even the same phrase tags within different syntactic contexts can vary greatly in size and internal structure, as the case of NP tags in Figure \ref{fig:figure1} shows. Here, the NP consisting of DT[the]-NN[stories] has a different internal structure than the NP consisting of NP[the film 's]-NNS[shortcomings].

One way of deriving \textit{structure-aware} tag representations from the original tag embeddings is to introduce a separate tag-level tree-LSTM which accepts the typical tag embeddings at each node of a tree and outputs the computed structure-aware tag representations for the nodes. Note that the module concentrates on extracting useful syntactic features by considering only the tags and structures of the trees, excluding word information. 

Formally, we denote a tag embedding for the tag attached to each node in a tree as $\textbf{e} \in\mathbb{R}^{d_\text{T}}$.
Then, the function of each cell in the tag tree-LSTM is defined in the following way.
Leaf nodes are defined by the following:
\begin{equation} \label{eq:4}
    \begin{bmatrix}
        \hat{\mathbf{c}} \\
        \hat{\mathbf{h}} \\
    \end{bmatrix}
    = \tanh{\left(\mathbf{U}_\text{T} \mathbf{e} + \mathbf{a}_\text{T}\right)}
\end{equation}

\noindent while non-leaf nodes are defined by the following:
\begin{equation} \label{eq:5}
    \begin{bmatrix}
        \hat{\mathbf{i}} \\
        \hat{\mathbf{f}}_l \\
        \hat{\mathbf{f}}_r \\ 
        \hat{\mathbf{o}} \\
        \hat{\mathbf{g}}
    \end{bmatrix}
    = 
    \begin{bmatrix}
         \sigma \\
         \sigma \\
         \sigma \\ 
         \sigma \\
         \tanh 
    \end{bmatrix}
    \Bigg( \mathbf{W_\text{T}}
    \begin{bmatrix}
        \hat{\mathbf{h}}_l \\
        \hat{\mathbf{h}}_r \\
        \mathbf{e} \\
    \end{bmatrix}
    + \mathbf{b}_\text{T} \Bigg)
\end{equation}
\begin{equation} \label{eq:6}
    \hat{\mathbf{c}} = \hat{\mathbf{f}}_l \odot \hat{\mathbf{c}}_l + \hat{\mathbf{f}}_r \odot \hat{\mathbf{c}}_r + \hat{\mathbf{i}} \odot \hat{\mathbf{g}}\\ 
\end{equation}
\begin{equation} \label{eq:7}
    \hat{\mathbf{h}} = \hat{\mathbf{o}} \odot \tanh{\left(\hat{\mathbf{c}}\right)}
\end{equation}
    
\noindent where $\hat{\mathbf{h}}, \hat{\mathbf{c}} \in\mathbb{R}^{d_\text{T}}$ represent the hidden state and cell state of each node in the tag tree-LSTM. 
We regard the hidden state ($\hat{\mathbf{h}}$) as a structure-aware tag representation for the node.
$ \mathbf{U}_\text{T} \in\mathbb{R}^{2d_\text{T} \times d_\text{T}}, \textbf{a}_\text{T} \in\mathbb{R}^{2d_\text{T}}, \mathbf{W}_\text{T} \in\mathbb{R}^{5d_\text{T} \times 3d_\text{T}}$, and $\mathbf{b}_\text{T} \in\mathbb{R}^{5d_\text{T}}$ are trainable parameters. 
The rest of the notation follows equations \ref{eq:1}, \ref{eq:2}, and \ref{eq:3}. 
In case of leaf nodes, the states are computed by a simple non-linear transformation.
Meanwhile, the composition function in a non-leaf node absorbs the tag embedding ($\mathbf{e}$) as an additional input as well as the hidden states of the two children nodes. The benefit of revising tag representations according to the internal structure is that the derived embedding is a function of the corresponding makeup of the node, rather than a monolithic, categorical tag.

With regard to the tags themselves, we conjecture that the taxonomy of the tags currently in use in many NLP systems is too complex to be utilized effectively in deep neural models, considering the specificity of many tag sets and the limited amount of data with which to train. Thus, we cluster POS (word-level) tags into 12 groups following the universal POS tagset \cite{petrov2012universal} and phrase-level tags into 11 groups according to criteria analogous to the case of words, resulting in 23 tag categories in total. In this work, we use the revised coarse-grained tags instead of the original ones. For more details, we refer readers to the supplemental materials.

\subsection{Leaf-LSTM}

An inherent shortcoming of RvNNs relative to sequential models is that each intermediate representation in a tree is unaware of its external context until all the information is gathered together at the root node. 
In other words, each composition process is prone to be locally optimized rather than globally optimized.

To mitigate this problem, we propose using a leaf-LSTM following the convention of some previous works \cite{eriguchi2016tree,yang2017towards,choi2018learning}, which is a typical LSTM that accepts a sequence of words in order. 
Instead of leveraging word embeddings directly, we can use each hidden state and cell state of the leaf-LSTM as input tokens for leaf nodes in a tree-LSTM, anticipating the proper contextualization of the input sequence. 

Formally, we denote a sequence of words in an input sentence as $w_{1:n}$ ($n$: the length of the sentence), and the corresponding word embeddings as $\mathbf{x}_{1:n}$. Then, the operation of the leaf-LSTM at time $t$ can be formulated as,
\begin{equation} \label{eq:8}
    \begin{bmatrix}
        \tilde{\mathbf{i}} \\
        \tilde{\mathbf{f}} \\
        \tilde{\mathbf{o}} \\
        \tilde{\mathbf{g}}
    \end{bmatrix}
    = 
    \begin{bmatrix}
         \sigma \\
         \sigma \\
         \sigma \\ 
         \tanh 
    \end{bmatrix}
    \Bigg( \mathbf{W}_\text{L}
    \begin{bmatrix}
        \tilde{\mathbf{h}}_{t-1} \\
        \mathbf{x}_t \\
    \end{bmatrix}
    + \mathbf{b}_\text{L} \Bigg)
\end{equation}
\begin{equation} \label{eq:9}
    \tilde{\mathbf{c}}_t = \tilde{\mathbf{f}} \odot \tilde{\mathbf{c}}_{t-1} + \tilde{\mathbf{i}} \odot \tilde{\mathbf{g}}\\ 
\end{equation}
\begin{equation} \label{eq:10}
    \tilde{\mathbf{h}}_t = \tilde{\mathbf{o}} \odot \tanh{\left(\tilde{\mathbf{c}}_t\right)}
\end{equation}

\noindent where $\mathbf{x}_t \in\mathbb{R}^{d_w}$ indicates an input word vector and $\tilde{\mathbf{h}}_t$, $\tilde{\mathbf{c}}_t \in\mathbb{R}^{d_h}$ represent the hidden and cell state of the LSTM at time $t$ ($\tilde{\mathbf{h}}_{t-1}$ corresponds to the hidden state at time $t$-1). $\mathbf{W}_\text{L}$ and $\mathbf{b}_\text{L} $ are learnable parameters. The remaining notation follows that of the tree-LSTM above. 

In experiments, we demonstrate that introducing a leaf-LSTM fares better at processing the input words of a tree-LSTM compared to using a feed-forward neural network. We also explore the possibility of its bidirectional setting in ablation study.

\subsection{SATA Tree-LSTM}

\begin{figure}[!t]
\centering
\includegraphics[width=1\columnwidth]{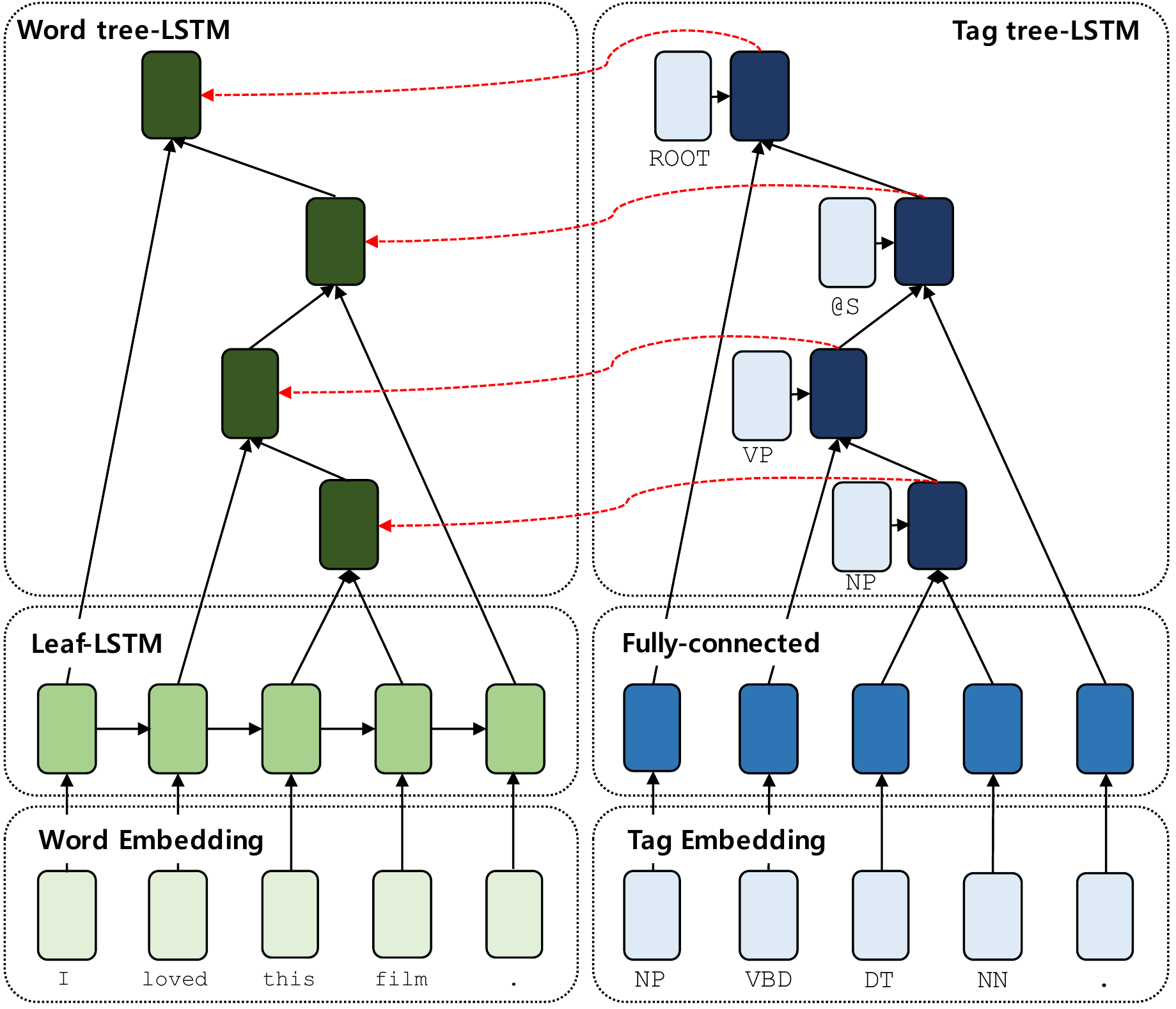}
\caption{A diagram of SATA Tree-LSTM. The model has two separate tree-LSTM modules, the right of which (tag tree-LSTM) extracts a structure-aware tag representation to control the composition function of the remaining tree-LSTM (word tree-LSTM). Fully-connected: one-layered non-linear transformation.}
\label{fig:figure2}
\end{figure}

In this section, we define \textbf{SATA Tree-LSTM} (\textbf{S}tructure-\textbf{A}ware \textbf{T}ag \textbf{A}ugmented \textbf{Tree-LSTM}, see Figure \ref{fig:figure2}) which joins a tag-level tree-LSTM (section 3.2), a leaf-LSTM (section 3.3), and the original word tree-LSTM together. 

As above we denote a sequence of words in an input sentence as $w_{1:n}$ and the corresponding word embeddings as $\mathbf{x}_{1:n}$. In addition, a tag embedding for the tag attached to each node in a tree is denoted by $\textbf{e} \in\mathbb{R}^{d_\text{T}}$. Then, we derive the final sentence representation for the input sentence with our model in two steps.

First, we compute structure-aware tag representations ($\hat{\mathbf{h}}$) for each node of a tree using the tag tree-LSTM (the right side of Figure \ref{fig:figure2}) as follows:
\begin{equation} \label{eq:11}
    \begin{bmatrix}
        \hat{\mathbf{c}} \\
        \hat{\mathbf{h}} \\
    \end{bmatrix}
    = 
    \begin{cases}
        \text{Tag-Tree-LSTM}(\mathbf{e}) & \text{if a leaf node} \\
        \text{Tag-Tree-LSTM}(\hat{\mathbf{h}}_l, \hat{\mathbf{h}}_r, \mathbf{e}) & \text{otherwise}
    \end{cases}
\end{equation}

\noindent where Tag-Tree-LSTM indicates the module we described in section 3.2. 

Second, we combine semantic units recursively on the word tree-LSTM in a bottom-up fashion. For leaf nodes, we leverage the Leaf-LSTM (the bottom-left of Figure \ref{fig:figure2}, explained in section 3.3) to compute $\tilde{\mathbf{c}}_{t}$ and $\tilde{\mathbf{h}}_{t}$ in sequential order, with the corresponding input $\mathbf{x}_t$.
\begin{equation} \label{eq:12}
    \begin{bmatrix}
        \tilde{\mathbf{c}}_{t} \\
        \tilde{\mathbf{h}}_{t} \\
    \end{bmatrix}
    = \text{Leaf-LSTM}(\tilde{\textbf{h}}_{t-1}, \textbf{x}_t)
\end{equation}

\noindent Then, the $\tilde{\mathbf{c}}_{t}$ and $\tilde{\mathbf{h}}_{t}$ can be utilized as input tokens to the word tree-LSTM, with the left (right) child of the target node corresponding to the $t$th word in the input sentence.
\begin{equation} \label{eq:13}
    \begin{bmatrix}
        \check{\textbf{c}}_{\{l, r\}} \\
        \check{\textbf{h}}_{\{l, r\}}
    \end{bmatrix}
    = 
    \begin{bmatrix}
        \tilde{\textbf{c}}_{t} \\
        \tilde{\textbf{h}}_{t}
    \end{bmatrix}
\end{equation}

In the non-leaf node case, we calculate phrase representations for each node 
in the word tree-LSTM (the upper-left of Figure \ref{fig:figure2}) recursively as follows:
\begin{equation} \label{eq:14}
    \check{\mathbf{g}} = \tanh{\left( \mathbf{U}_\text{w} 
    \begin{bmatrix}
        \check{\mathbf{h}}_l \\
        \check{\mathbf{h}}_r \\
    \end{bmatrix}
    + \mathbf{a}_\text{w} \right)}
\end{equation}
\begin{equation} \label{eq:15}
    \begin{bmatrix}
        \check{\mathbf{i}} \\
        \check{\mathbf{f}}_l \\
        \check{\mathbf{f}}_r \\ 
        \check{\mathbf{o}}
    \end{bmatrix}
    = 
    \begin{bmatrix}
         \sigma \\
         \sigma \\
         \sigma \\ 
         \sigma
    \end{bmatrix}
    \Bigg( \mathbf{W_\text{w}}
    \begin{bmatrix}
        \check{\mathbf{h}}_l \\
        \check{\mathbf{h}}_r \\
        \hat{\mathbf{h}} \\
    \end{bmatrix}
    + \mathbf{b}_\text{w} \Bigg)
\end{equation}
\begin{equation} \label{eq:16}
    \check{\mathbf{c}} = \check{\mathbf{f}}_l \odot \check{\mathbf{c}}_l + \check{\mathbf{f}}_r \odot \check{\mathbf{c}}_r + \check{\mathbf{i}} \odot \check{\mathbf{g}}
\end{equation}
\begin{equation} \label{eq:17}
    \check{\mathbf{h}} = \check{\mathbf{o}} \odot \tanh{\left(\check{\mathbf{c}}\right)}
\end{equation}

\noindent where $\check{\mathbf{h}}$, $\check{\mathbf{c}} \in \mathbb{R}^{d_h}$ represent the hidden and cell state of each node in the word tree-LSTM. $\mathbf{U}_\text{w} \in \mathbb{R}^{d_h \times 2d_h}$, $\mathbf{W}_\text{w} \in \mathbb{R}^{4d_h \times \left(2d_h+d_\text{T}\right)}$, $\mathbf{a}_\text{w} \in \mathbb{R}^{d_h}$, $\mathbf{b}_\text{w} \in \mathbb{R}^{4d_h}$ are learned parameters. The remaining notation follows those of the previous sections.
Note that the structure-aware tag representations ($\hat{\mathbf{h}}$) are only utilized to control the gate functions of the word tree-LSTM in the form of additional inputs, and are not involved in the semantic composition ($\check{\mathbf{g}}$) directly.

Finally, the hidden state of the root node ($\check{\mathbf{h}}_\text{root}$) in the word-level tree-LSTM becomes the final sentence representation of the input sentence.

\section{Experiment and Discussion}

\subsection{Quantitative Analysis}

\subsubsection{Sentence classification tasks}

One of the most basic approaches to evaluate a sentence encoder is to measure the classification performance with the sentence representations made by the encoder. Thus, we conduct experiments on the following five datasets. (Summary statistics for the datasets are reported in the supplemental materials.)

\begin{itemize}
    \item \textbf{MR}: A group of movie reviews with binary (positive / negative) classes. \cite{pang2005seeing}
    \item \textbf{SST-2}: Stanford Sentiment Treebank \cite{socher2013recursive}. 
    Similar to MR, but each review is provided in the form of a binary parse tree whose nodes are annotated with numeric sentiment values. 
    For SST-2, we only consider binary (positive / negative) classes.
    \item \textbf{SST-5}: Identical to SST-2, but the reviews are grouped into fine-grained (very negative, negative, neutral, positive, very positive) classes.
    \item \textbf{SUBJ}: Sentences grouped as being either subjective or objective (binary classes). \cite{pang2004sentimental}
    \item \textbf{TREC}: A dataset which groups questions into six different question types (classes). \cite{li2002learning}
\end{itemize}

As a preprocessing step, we construct parse trees for the sentences in the datasets using the Stanford PCFG parser \cite{klein2003accurate}.
Because syntactic tags are by-products of constituency parsing, we do not need further preprocessing. 

To classify the sentence given our sentence representation ($\check{\mathbf{h}}_\text{root}$), we use one fully-connected layer with a ReLU activation, followed by a softmax classifier. The final predicted probability distribution of the class $y$ given the sentence $w_{1:n}$ is defined as follows,
\begin{equation}
    \mathbf{s} = \text{ReLU}(\mathbf{W}_\text{s} \check{\mathbf{h}}_\text{root}+ \mathbf{b}_\text{s})
\end{equation}
\begin{equation}
    p(y|w_{1:n}) = \text{softmax}(\mathbf{W}_\text{c}\mathbf{s} + \mathbf{b}_\text{c})
\end{equation}

\noindent where $\textbf{s} \in \mathbb{R}^{d_\text{s}}$ is the computed task-specific sentence representation for the classifier, and $\textbf{W}_\text{s} \in \mathbb{R}^{d_\text{s} \times d_h}$, $\textbf{W}_\text{c} \in \mathbb{R}^{d_\text{c} \times d_s}$, $\textbf{b}_\text{s} \in \mathbb{R}^{d_s}$, $\textbf{b}_\text{c} \in \mathbb{R}^{d_c}$ are trainable parameters. As an objective function, we use the cross entropy of the predicted and true class distributions.

The results of the experiments on the five datasets are shown in table \ref{table1}. 
In this table, we report the test accuracy of our model and various other models on each dataset in terms of percentage.
To consider the effects of random initialization, we report the best numbers obtained from each several runs with hyper-parameters fixed.

Compared with the previous syntactic tree-based models as well as other neural models, our SATA Tree-LSTM shows superior or competitive performance on all tasks.
Specifically, our model achieves new state-of-the-art results within the tree-structured model class on 4 out of 5 sentence classification tasks---SST-2, SST-5, MR, and TREC. 
The model shows its strength, in particular, when the datasets provide phrase-level supervision to facilitate tree structure learning (i.e. SST-2, SST-5).
Moreover, the numbers we report for SST-5 and TREC are competitive to the existing state-of-the-art results including ones from structurally pre-trained models such as ELMo \cite{peters2018deep}, proving our model's superiority.
Note that the SATA Tree-LSTM also outperforms the recent latent tree-based model, indicating that modeling a neural model with explicit linguistic knowledge can be an attractive option.

On the other hand, a remaining concern is that our SATA Tree-LSTM is not robust to random seeds when the size of a dataset is relatively small, as tag embeddings are randomly initialized rather than relying on pre-trained ones in contrast with the case of words. 
From this observation, we could find out there needs a direction of research towards pre-trained tag embeddings.

\begin{table*}[t!]
\centering
\resizebox{0.7\textwidth}{!}{
\begin{tabular}{|l|c|c|c|c|c|}
\hline
\multicolumn{1}{|c|}{\textbf{Models}} & \textbf{SST-2} & \textbf{SST-5} & \textbf{MR} & \textbf{SUBJ} & \textbf{TREC} \\
\hline \hline
\multicolumn{6}{|l|}{\textbf{Tree-structured models}} \\
\hline
RNTN \cite{socher2013recursive} & 85.4 & 45.7 & - & - & - \\
AdaMC-RNTN \cite{dong2014adaptive} & 88.5 & 46.7 & - & - & - \\
TE-RNTN \cite{qian2015learning} & 87.7 & 49.8 & - & - & - \\
TBCNN \cite{mou2015discriminative} & 87.9 & 51.4 & - & - & 96.0 \\
Tree-LSTM \cite{tai2015improved} & 88.0 & 51.0 & - & - & - \\
AdaHT-LSTM-CM \cite{liu2017adaptive} & 87.8 & 50.2 & 81.9 & 94.1 & - \\
DC-TreeLSTM \cite{liu2017dynamic} & 87.8 & - & 81.7 & 93.7 & 93.8 \\
TE-LSTM \cite{huang2017encoding} & 89.6 & 52.6 & 82.2 & - & - \\
BiConTree \cite{teng2017head} & 90.3 & 53.5 & - & - & 94.8 \\
Gumbel Tree-LSTM$^\star$ \cite{choi2018learning} & 90.7 & 53.7 & - & - & - \\ 
TreeNet \cite{cheng2018treenet} & - & - & 83.6 & \underline{95.9} & 96.1 \\
\textbf{SATA Tree-LSTM (Ours)} & \textbf{91.3} & \underline{\textbf{54.4}} & \textbf{83.8} & \textbf{95.4} & \underline{\textbf{96.2}} \\
\hline \hline
\multicolumn{6}{|l|}{\textbf{Other neural models}} \\
\hline
CNN \cite{kim2014convolutional} & 88.1 & 48.0 & 81.5 & 93.4 & 93.6 \\
AdaSent \cite{zhao2015self} & - & - & 83.1 & 95.5 & 92.4 \\
LSTM-CNN \cite{zhou2016text} & 89.5 & 52.4 & 82.3 & 94.0 & 96.1 \\
byte-mLSTM$^\dagger$ \cite{radford2017learning} & \underline{91.8} & 52.9 & \underline{86.9} & 94.6 & - \\
BCN + Char + CoVe$^\dagger$ \cite{mccann2017learned} & 90.3 & 53.7 & - & - & 95.8 \\
BCN + Char + ELMo$^\dagger$ \cite{peters2018deep} & - & \underline{54.7$\pm$0.5} & - & - & - \\
\hline
\end{tabular}
}
\caption{The comparison of various models on different sentence classification tasks.  We report the test accuracy of each model in percentage. Our SATA Tree-LSTM shows superior or competitive performance on all tasks, compared to previous tree-structured models as well as other sophisticated models. 
$\star$: Latent tree-structured models. $\dagger$: Models which are pre-trained with large external corpora.}
\label{table1}
\end{table*}

\subsubsection{Natural language inference}

To estimate the performance of our model beyond the tasks requiring only one sentence at a time, we conduct an experiment on the Stanford Natural Language Inference \cite{snli} dataset, each example of which consists of two sentences, the premise and the hypothesis. Our objective given the data is to predict the correct relationship between the two sentences among three options--- contradiction, neutral, or entailment.

We use the siamese architecture to encode both the premise ($p_{1:m}$) and hypothesis ($h_{1:n}$) following the standard of sentence-encoding models in the literature. (Specifically, $p_{1:m}$ is encoded as $\check{\mathbf{h}}_\text{root}^p \in \mathbb{R}^{d_h}$ and $h_{1:n}$ is encoded as $\check{\mathbf{h}}_\text{root}^h \in \mathbb{R}^{d_h}$ with the same encoder.) Then, we leverage some heuristics \cite{mou2016natural}, followed by one fully-connected layer with a ReLU activation and a softmax classifier. Specifically, 
\begin{equation}
    \mathbf{z} = \left[ \check{\mathbf{h}}_\text{root}^p; \check{\mathbf{h}}_\text{root}^h; | \check{\mathbf{h}}_\text{root}^p - \check{\mathbf{h}}_\text{root}^h |; \check{\mathbf{h}}_\text{root}^p \odot \check{\mathbf{h}}_\text{root}^h \right]
\end{equation}
\begin{equation}
    \mathbf{s} = \text{ReLU}(\mathbf{W}_\text{s} \mathbf{z} + \mathbf{b}_\text{s})
\end{equation}
\begin{equation}
    p(y|p_{1:m}, h_{1:n}) = \text{softmax}(\mathbf{W}_\text{c}\textbf{s} + \mathbf{b}_\text{c})
\end{equation}

\noindent where $\textbf{z} \in \mathbb{R}^{4d_h}$, $\textbf{s} \in \mathbb{R}^{d_s}$ are intermediate features for the classifier and
$\textbf{W}_\text{s} \in \mathbb{R}^{d_\text{s} \times 4d_h}$, $\textbf{W}_\text{c} \in \mathbb{R}^{d_\text{c} \times d_s}$, $\textbf{b}_\text{s} \in \mathbb{R}^{d_s}$, $\textbf{b}_\text{c} \in \mathbb{R}^{d_c}$ are again trainable parameters.

Our experimental results on the SNLI dataset are shown in table \ref{table2}. In this table, we report the test accuracy and number of trainable parameters for each model. Our SATA-LSTM again demonstrates its decent performance compared against the neural models built on both syntactic trees and latent trees, as well as the non-tree models.
(Latent Syntax Tree-LSTM: \citeauthor{yogatama2017learning} (\citeyear{yogatama2017learning}), 
Tree-based CNN: \citeauthor{mou2016natural} (\citeyear{mou2016natural}), 
Gumbel Tree-LSTM: \citeauthor{choi2018learning} (\citeyear{choi2018learning}),   
NSE: \citeauthor{munkhdalai2017neural} (\citeyear{munkhdalai2017neural}),   
Reinforced Self-Attention Network: \citeauthor{shen2018reinforced} (\citeyear{shen2018reinforced}), 
Residual stacked encoders: \citeauthor{nie2017shortcut} (\citeyear{nie2017shortcut}), 
BiLSTM with generalized pooling: \citeauthor{chen2018enhancing} (\citeyear{chen2018enhancing}).)
Note that the number of learned parameters in our model is also comparable to other sophisticated models, showing the efficiency of our model. 

Even though our model has proven its mettle, the effect of tag information seems relatively weak in the case of SNLI, which contains a large amount of data compared to the others.
One possible explanation is that neural models may learn some syntactic rules from large amounts of text when the text size is large enough, reducing the necessity of external linguistic knowledge.
We leave the exploration of the effectiveness of tags relative to data size for future work.

\subsubsection{Experimental details}

Here we go over the settings common across our models during experimentation. For more task-specific details, refer to the supplemental materials.

For our input embeddings, we used 300 dimensional 840B GloVe \cite{pennington2014glove} as pre-trained word embeddings, and tag representations were randomly sampled from the uniform distribution [-0.005, 0.005]. Tag vectors are revised during training while the fine-tuning of the word embedding depends on the task. 
Our models were trained using the Adam \cite{kingma2014adam} or Adadelta \cite{zeiler2012adadelta} optimizer, depending on task. 
For regularization, weight decay is added to the loss function except for SNLI following \citeauthor{loshchilov2017fixing} (\citeyear{loshchilov2017fixing}) and Dropout \cite{srivastava2014dropout} is also applied for the word embeddings and task-specific classifiers.
Moreover, batch normalization \cite{ioffe2015batch} is adopted for the classifiers.
As a default, all the weights in the model are initialized following \citeauthor{he2015delving} (\citeyear{he2015delving}) and the biases are set to 0.
The total norm of the gradients of the parameters is clipped not to be over 5 during training.

Our best models for each dataset were chosen by validation accuracy in cases where a validation set was provided as a part of the dataset. Otherwise, we perform a grid search on probable hyper-parameter settings, or run 10-fold cross-validation in cases where even a test set does not exist.

\begin{table}[t!]
\centering
\small
\resizebox{0.9\columnwidth}{!}{
\begin{tabular}{|l|c|c|}
\hline
\multicolumn{1}{|c|}{\textbf{Models}} & \textbf{Acc.} & \textbf{\# Params} \\
\hline \hline
\multicolumn{3}{|l|}{\textbf{Tree-structured models}} \\
\hline
100D Latent Syntax Tree-LSTM$^\star$ & 80.5 & 500K \\
300D Tree-based CNN & 82.1 & 3.5M \\
300D SPINN-PI & 83.2 & 3.7M \\ 
300D Gumbel Tree-LSTM$^\star$ & 85.6 & 2.9M \\
\textbf{300D SATA Tree-LSTM (Ours)} & \textbf{85.9} & \textbf{3.3M} \\
\hline \hline
\multicolumn{3}{|l|}{\textbf{Other neural models}} \\
\hline
300D NSE & 84.6 & 3.0M \\
300D Reinforced Self-Attention Network & 86.3 & 3.1M \\
600D Residual stacked encoders & 86.0 & 29M \\
600D BiLSTM with generalized pooling & \underline{86.6} & 65M \\
\hline
\end{tabular}}
\caption{The accuracy of diverse models on Stanford Natural Language Inference. For fair comparison, we only consider sentence-encoding based models. Our model achieves a comparable result with a moderate number of parameters. 
$\star$: Latent tree models.}
\label{table2}
\end{table}

\subsection{Ablation Study}

In this section, we design an ablation study on the core modules of our model to explore their effectiveness. 
The dataset used in this experiment is SST-2. 
To conduct the experiment, we only replace the target module with other candidates while maintaining the other settings.
To be specific, we focus on two modules, the leaf-LSTM and structure-aware tag embeddings (tag-level tree-LSTM). 
In the first case, the leaf-LSTM is replaced with a fully-connected layer with a $\tanh$ activation or Bi-LSTM. 
In the second case, we replace the structure-aware tag embeddings with naive tag embeddings or do not employ them at all.

\begin{figure}[!t]
\centering
\includegraphics[width=0.9\columnwidth]{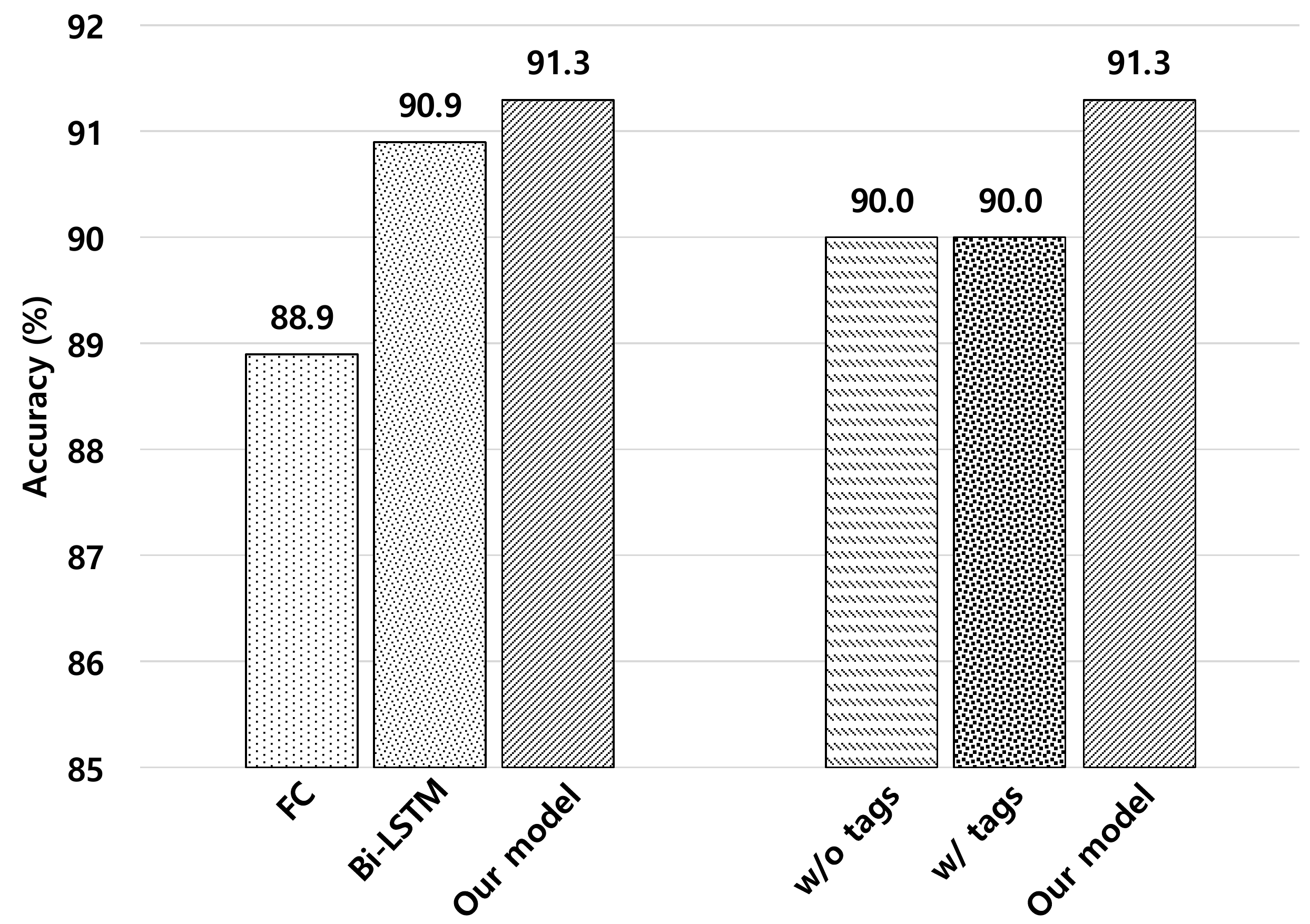}
\caption{An ablation study on the core modules of our model. The test accuracy of each model on SST-2 is reported. The results demonstrate that the modules play an important role for achieving the superior performance of our model. FC: A fully connected-layer with a $\tanh$ function. w/o tags: Tag embeddings are not used. w/ tags: The naive tag embeddings are directly inserted into each node of a tree.}
\label{fig:figure3}
\end{figure}

The experimental results are depicted in Figure \ref{fig:figure3}. 
As the chart shows, our model outperforms all the other options we have considered. 
In detail, the left part of the chart shows that the leaf-LSTM is the most effective option compared to its competitors. 
Note that the sequential leaf-LSTM is somewhat superior or competitive than the bidirectional leaf-LSTM when both have a comparable number of parameters.
We conjecture this may because a backward LSTM does not add additional useful knowledge when the structure of a sentence is already known.
In conclusion, we use the uni-directional LSTM as a leaf module because of its simplicity and remarkable performance.

Meanwhile, the right part of the figure demonstrates that our newly introduced structure-aware embeddings have a real impact on improving the model performance. 
Interestingly, employing the naive tag embeddings made no difference in terms of the test accuracy, even though the absolute validation accuracy increased (not reported in the figure). 
This result supports our assumption that tag information should be considered in the structure.

\subsection{Qualitative Analysis}

In previous sections, we have numerically demonstrated that our model is effective in encouraging useful composition of semantic units. Here, we directly investigate the computed representations for each node of a tree, showing that the remarkable performance of our model is mainly due to the gradual and recursive composition of the intermediate representations on the syntactic structure.

\begin{figure}[!t]
\centering
\includegraphics[width=0.95\columnwidth]{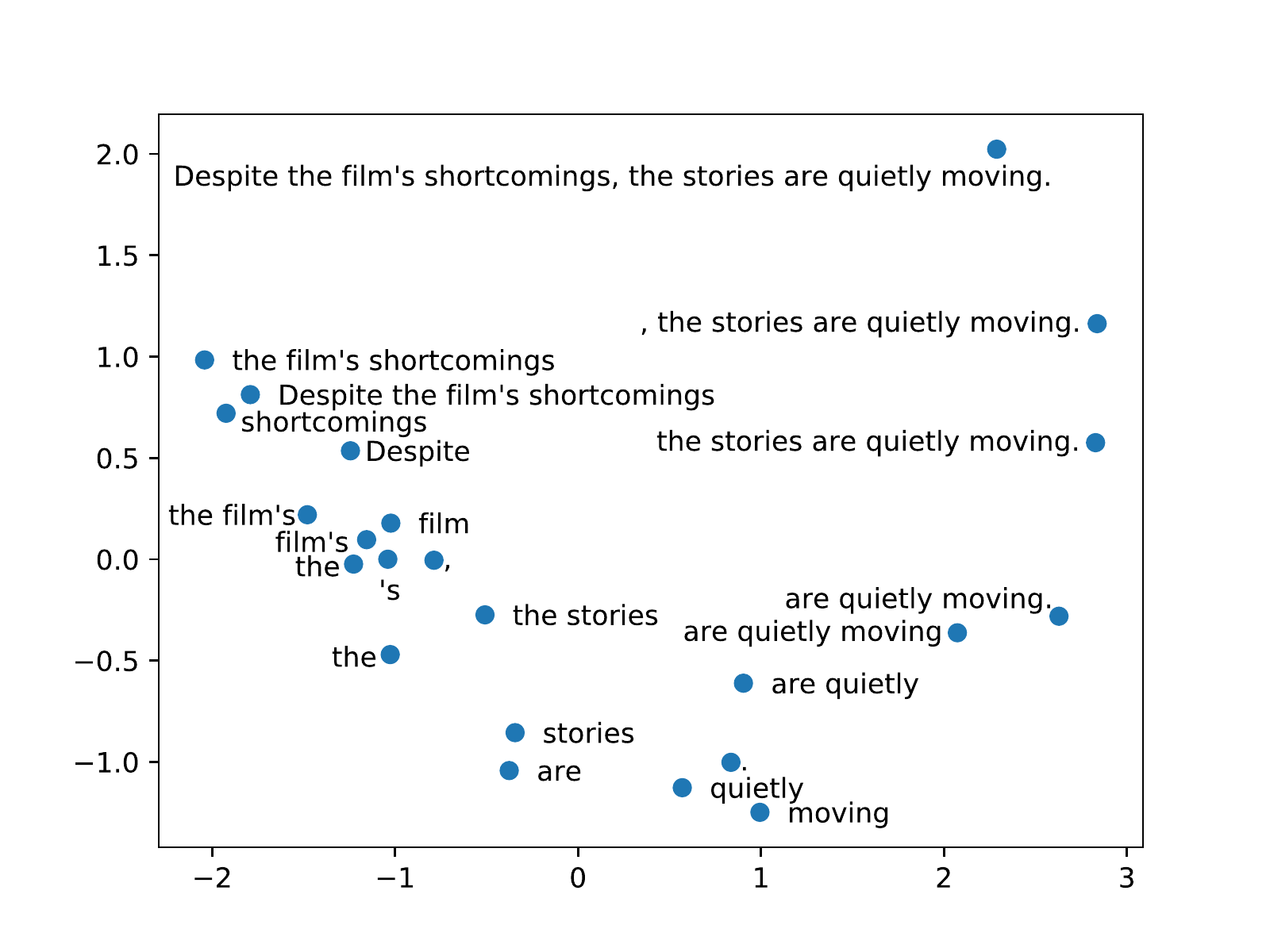}
\caption{A scatter plot whose points represent the intermediate representations for each node of the tree in Figure 1. From this figure, we can see the tendency of constructing the representations recursively from the low to the high level.}
\label{fig:figure4}
\end{figure}

To observe the phrase-level embeddings at a glance, we draw a scatter plot in which a point represents the corresponding intermediate representation. 
We utilize PCA (Principal Component Analysis) to project the representations into a two-dimensional vector space. 
As a target parse tree, we reuse the one seen in Figure \ref{fig:figure1}. 
The result is shown in Figure \ref{fig:figure4}.

From this figure, we confirm that the intermediate representations have a hierarchy in the semantic space, which is very similar to that of the parse tree. 
In other words, as many tree-structured models pursue, we can see the tendency of constructing the representations from the low-level (the bottom of the figure) to the high-level (the top-left and top-right of the figure), integrating the meaning of the constituents recursively.
An interesting thing to note is that the final sentence representation is near that of the phrase \textit{`, the stories are quietly moving.'} rather than that of \textit{`Despite the film's shortcomings'}, catching the main meaning of the sentence.

\section{Conclusion}

We have proposed a novel RvNN architecture to fully utilize linguistic priors.
A newly introduced tag-level tree-LSTM demonstrates that it can effectively control the composition function of the corresponding word-level tree-LSTM. 
In addition, the proper contextualization of the input word vectors results in significant performance improvements on several sentence-level tasks.
For future work, we plan to explore a new way of exploiting dependency trees effectively, similar to the case of constituency trees.

\section*{Acknowledgments}
We thank anonymous reviewers for their constructive and fruitful comments. This work was supported by the National Research Foundation of Korea (NRF) grant
funded by the Korea government (MSIT) (NRF2016M3C4A7952587).

\fontsize{9pt}{10pt} \selectfont
\bibliography{aaai2019}
\bibliographystyle{aaai}

\end{document}